\documentclass{article}

\usepackage[preprint]{neurips_2026}

\usepackage[utf8]{inputenc} % allow utf-8 input
\usepackage[T1]{fontenc}    % use 8-bit T1 fonts
\usepackage{hyperref}       % hyperlinks
\usepackage{url}            % simple URL typesetting
\usepackage{booktabs}       % professional-quality tables
\usepackage{amsfonts}       % blackboard math symbols
\usepackage{nicefrac}       % compact symbols for 1/2, etc.
\usepackage{microtype}      % microtypography
\usepackage{xcolor}         % colors
\usepackage{booktabs}       % tables
\usepackage{multirow}
\usepackage{makecell}
\usepackage{graphicx}
\usepackage{wrapfig}
\usepackage[capitalize]{cleveref}
\crefname{equation}{Eq.}{Eqs.}
\Crefname{equation}{Eq.}{Eqs.}
\crefname{table}{Table}{Tables}
\Crefname{table}{Table}{Tables}
\usepackage{algorithm}
\usepackage{algorithmic}

\usepackage{caption}
\usepackage{amsmath, amssymb}
\usepackage{enumitem}
\usepackage{tabularx}
\usepackage{array}

\title{\mymethod: Layer-wise Token Pruning for Omni-modal LLMs via Query-Guidance}

\author{Yeo Jeong Park \quad Hyemi Jang \quad Minseo Choi \\ 
\textbf{Jongsun Lee \quad Jooyoung Choi \quad Yongkweon Jeon\thanks{Corresponding Author}} \\
  Samsung Research\\
  \texttt{\{yeo\_j.park, hye\_mi.jang, mins.choi\}@samsung.com} \\
  \texttt{\{jongsun3.lee, jy9718.choi, dragwon.jeon\}@samsung.com} \\ 
}

\newcommand{\mymethod}{OmniDrop}

\begin{document}

\maketitle
\begin{abstract}
Omni-modal large language models have demonstrated remarkable potential in holistic multimodal understanding; however, the token explosion caused by high-resolution audio and video inputs remains a critical bottleneck for real-time applications and long-form reasoning.
Existing omni-modal token compression methods typically prune tokens at the input embedding level, relying on audio-video similarity or temporal co-occurrence as proxies for semantic relevance. 
In practice, such assumptions are often unreliable.
To address this limitation, we propose \mymethod, a \textit{training-free, layer-wise token pruning} framework that progressively prunes audiovisual tokens within the LLM decoder layers rather than at the input-level, allowing early layers to preserve sufficient omni-modal information fusion before aggressively removing tokens in deeper layers.
We further utilize \textit{text queries as guidance} for modality-agnostic and task-adaptive token pruning. 
We also introduce a \textit{temporal diversity score} that encourages balanced token survival to preserve global temporal context. 
Experimental results across various audiovisual benchmarks demonstrate that \mymethod~outperforms all baselines by up to 3.58 points while reducing prefill latency by up to 40\% and memory usage by up to 14.7\%.
\end{abstract}

\section{Introduction}

The recent paradigm shift in large language models (LLMs) has led to the emergence of Omni-modal LLMs (Omni-LLMs)~\cite{gemini,vita,gpt-4o,baichuan,qwen,minicpm}, which seamlessly integrate and process multiple modalities such as text, audio, and video. By enabling a more holistic understanding of complex environments, Omni-LLMs have demonstrated significant potential in diverse applications, ranging from real-time multimodal assistants to sophisticated video reasoning tasks~\cite{omniagent,apagent,agent-omni,nemotron}. Despite their impressive capabilities, a critical bottleneck remains: computational efficiency.

A primary challenge in Omni-LLMs is the \textit{token explosion} caused by high-resolution temporal data.
Audio and video signals produce significantly more tokens than text~\cite{song2024moviechat}; for instance, a mere one-minute video clip is projected into more than 10k audio and video tokens by encoders in Qwen2.5-Omni~\cite{qwen}. 
This massive sequence length leads to quadratic growth in both computational and memory overheads~\cite{tay2022efficient,vaswani2017attention}, hindering real-time interactions and limiting the model's scalability for long-form content.

Recent studies have introduced various token compression methods to address these inefficiencies, including audio-guided compression~\cite{omnizip}, modality-asymmetric compression~\cite{omnisift}, and semantic-aware alignment~\cite{dash}. Despite their methodological diversity, these methods share a common design choice: they perform compression at the input embedding level, 
assuming that audio-video tokens are semantically aligned either through a shared representation space or through temporal co-occurrence.

However, these assumptions are often unreliable for two reasons. First, audio and video embeddings may not be sufficiently aligned in the shared input embedding space. Although audio and video tokens are projected into the same LLM embedding space, they occupy distinct subspaces, suggesting limited cross-modal alignment before LLM processing. Second, temporal co-occurrence does not necessarily imply semantic correspondence. An audio event and a visual event may occur at the same time but describe different aspects of the scene or even originate from different sources. For example, background music, off-screen narration, ambient sounds, and scene transitions may carry information that is not directly grounded in the visible content. As a result, input-level pruning~\cite{omnizip, dash} that relies on embedding-level distance between audio and video tokens or on temporal co-occurrence may discard important modality-specific evidence and lead to suboptimal token selection.
In contrast, we empirically observe that audio and video tokens become progressively integrated through the LLM decoder layers. As the fusion of omni-modal representations progresses, the model begins to distinguish informative tokens from the middle layers onward.

We propose \mymethod, 
a training-free, layer-wise token pruning framework built on three key ideas.
First, we prune audiovisual tokens progressively across decoder layers rather than at the input stage.
This design allows shallow layers to first exchange omni-modal information before redundant tokens are removed in deeper layers. 
Second, we use the text query as a semantic mediator between modalities. Instead of directly comparing audio and video tokens, we evaluate how each modality responds to the text query.
This enables dynamically preserving more audio tokens for audio-centric tasks and more visual tokens for video-centric tasks, ensuring minimal information loss during compression.
Third, due to query-guided pruning, tokens from a few highly responsive tokens may dominate, while other tokens are wiped out.
To counteract this, we introduce a temporal diversity score (TDS) built upon the chunked prefilling mechanism commonly used in Omni-LLMs, encouraging balanced token survival across temporal segments.

Experimental results on diverse audiovisual benchmarks~\cite{videomme,worldsense,avut,shortvid} show that our method consistently outperforms existing omni-modal token compression approaches. Our framework achieves superior efficiency-performance trade-offs, offering a practical solution for scalable and responsive Omni-LLMs.

Our contributions are summarized as follows:
\begin{itemize}[leftmargin=*]
    \item We analyze omni-modal processing in Omni-LLMs and show that input-level audio-video similarity provides a weak signal for token pruning, as audio and video embeddings are not sufficiently aligned before LLM processing. 
    \item We propose \mymethod, a training-free, layer-wise token pruning framework for Omni-LLMs. It adopts a progressive pruning schedule that leverages text queries to estimate query-conditioned audio/video token relevance, enabling task-adaptive token pruning without training or architectural modifications.
    \item We validate \mymethod~through extensive experiments on diverse audiovisual understanding benchmarks. Our method aggressively reduces the token retention ratio to as low as 20\%, achieving up to 3.58-point gains over other baselines while reducing prefill latency by up to 40\% and memory usage by up to 14.7\%.
\end{itemize}

\section{Preliminary \& Motivation}

\subsection{Architecture of Omni-LLMs}
Typically, an Omni-LLM consists of multimodal encoders---an audio encoder $\Phi_A$ and a video encoder $\Phi_V$---integrated with an LLM backbone. Given audio data $\mathcal{A}$ and video data $\mathcal{V}$, the encoders project the multimodal inputs into the joint embedding space of the LLM:
\begin{equation}
Z_A = \Phi_A(\mathcal{A}) \in \mathbb{R}^{n_A \times d}, \quad Z_V = \Phi_V(\mathcal{V}) \in \mathbb{R}^{n_V \times d},    
\end{equation}
where $n_A$ and $n_V$ denote the number of audio and video embedding tokens, respectively, and $d$ represents the hidden dimension of the LLM. These projected embeddings $Z_A$ and $Z_V$ are then injected into the LLM in a manner analogous to text tokens.

To ensure temporal alignment between the visual and auditory streams, Omni-LLMs employ a time-interleaving method. The audiovisual sequence is segmented into several chunks based on a fixed temporal window (\textit{e.g.}, 2 seconds). Within each chunk $i$, the corresponding video representations $V_i\in \mathbb{R}^{n_v \times d}$ and audio representations $A_i\in \mathbb{R}^{n_a \times d}$ are arranged sequentially, where $n_v$ and $n_a$ represent the number of video and audio tokens per chunk. This structured arrangement ensures that tokens from both modalities share the same temporal context are processed simultaneously. Consequently, the final input sequence $\mathbf{X}$ fed into the LLM is organized as a concatenated stream:
\begin{equation}
\mathbf{X} = [ \text{T}_{sys}, V_1, A_1, V_2, A_2, \dots, V_m, A_m, \text{T}_{query} ],
\end{equation}
where $\text{T}_{sys}$ is the system prompt, $m$ is the total number of chunks and $\text{T}_{query}$ represents the user’s instructions or query prompts. 

\begin{figure}[t]
  \centering
  \includegraphics[width=\textwidth]{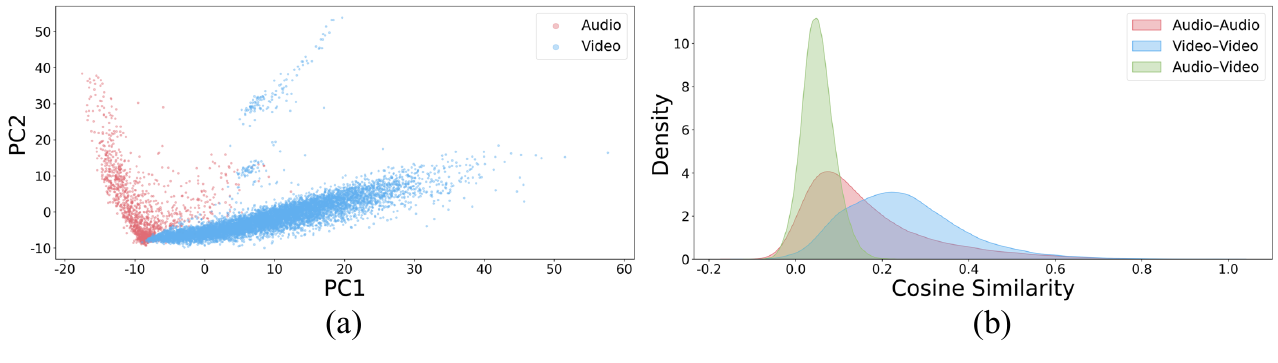}
  \caption{(a) PCA visualization of audio and video token distributions in the LLM embedding space. (b) Distribution of cosine similarities among audio-audio, video-video and audio-video token pairs.}
  \label{fig:cosine_sim}
\end{figure}

\subsection{Revisiting the utility of embedding-level cross-modal similarity}
\label{sec:revisit}

Recent token compression methods~\cite{omnizip,dash,echoing} for Omni-LLMs rely on audio-video correspondence at the input embedding level. However, this signal might be unreliable, and we provide multiple evidence below.

\textbf{Pre-LLM embeddings are weakly aligned across modalities.}

\begin{wraptable}{r}{0.40\linewidth}
    \vspace{-10pt}
    \scriptsize
    \caption{Ablation studies on the OmniZip~\cite{omnizip} at 35\% retention ratio on WorldSense~\cite{worldsense} benchmark.}
    \centering
    \label{tab:omnizip}
    \small
    \setlength{\tabcolsep}{4pt}
    \begin{tabular}{lcc}
    \toprule
    \multirow{2}{*}{Method} & \multicolumn{2}{c}{WS Acc. (\%)} \\
    \cmidrule{2-3}
    & 3B & 7B \\
    \midrule
    OmniZip & 46.15 & 46.37 \\
    \quad w/o AC stage & 46.19 & \textbf{46.78} \\
    \quad Random-$k$ sampling & \textbf{46.25} & 46.50 \\
    \bottomrule
    \end{tabular}
    \vspace{-10pt}
\end{wraptable}

PCA visualization in Fig.~\ref{fig:cosine_sim} (a) shows that audio and video tokens tend to occupy distinct subspaces rather than forming a semantically shared representation. 
Fig.~\ref{fig:cosine_sim} (b) shows the distribution of cosine similarities between intra- and cross-modal tokens. 
Intra-modal similarities exhibit broad distributions, whereas cross-modal ones range from 0.0 to 0.2, precluding the identification of meaningful relationships.  
Hence, cross-modal cosine similarity at the input embedding level may be uninformative for identifying semantically aligned cross-modal token pairs.

\textbf{Similarity-based selection is not better than random.}
In OmniZip~\cite{omnizip}, audio anchor consolidation (AC) merges non-salient audio tokens into anchors based on cross-modal similarity. 
This process relies on the assumption that input-level similarity signifies semantic alignment.

To test this assumption, we conduct an ablation study with a retained ratio of 35\% on WorldSense~\cite{worldsense} benchmark, comparing the original AC against two variants: (1) entirely removing AC stage, and (2) selecting merged tokens via random-k sampling instead of top-k selection by similarities.

As shown in Tab.~\ref{tab:omnizip}, removing the AC stage or employing random-$k$ sampling yields better or comparable results compared to the original similarity-based AC across all configurations. 
For instance, removing AC improve the WorldSense~\cite{worldsense} score for OmniZip-7B and OmniZip-3B to 46.78\% and 46.19\%, respectively. Even random-$k$ sampling shows a slight performance gain over similarity-based selection (\textit{e.g.}, 46.25\% vs. 46.15\% for the 3B model). These observations reinforce the notion that cosine similarity at the embedding level provides no distinct advantage over random token selection.

Given the insufficient alignment of pre-LLM embeddings, cross-modal similarity-based compression at the input level remains suboptimal. 
Instead, we perform pruning within intermediate decoder layers, leveraging emergent omni-modal interactions for more effective token selection.

\subsection{Analysis of omni-modal processing in LLM decoder layers}
\label{sec:analysis}

\textbf{Token importance is query-dependent.}
To better understand how audio and video tokens are processed inside the LLM, we analyze text-to-audiovisual attention patterns across decoder layers in Qwen2.5-Omni-7B. In particular, we investigate how the model responds to different text queries when the same audio-video input is provided. Given two queries, \textit{``Who said in the video that `...'?''} and \textit{``In which part of the video does the man with ... appear?''}, we visualize the attention scores from text tokens to audiovisual tokens at each decoder layer. For clearer visualization, we average the attention scores of tokens within each audiovisual chunk.

\begin{figure}[t]
  \centering\includegraphics[width=0.9\textwidth]{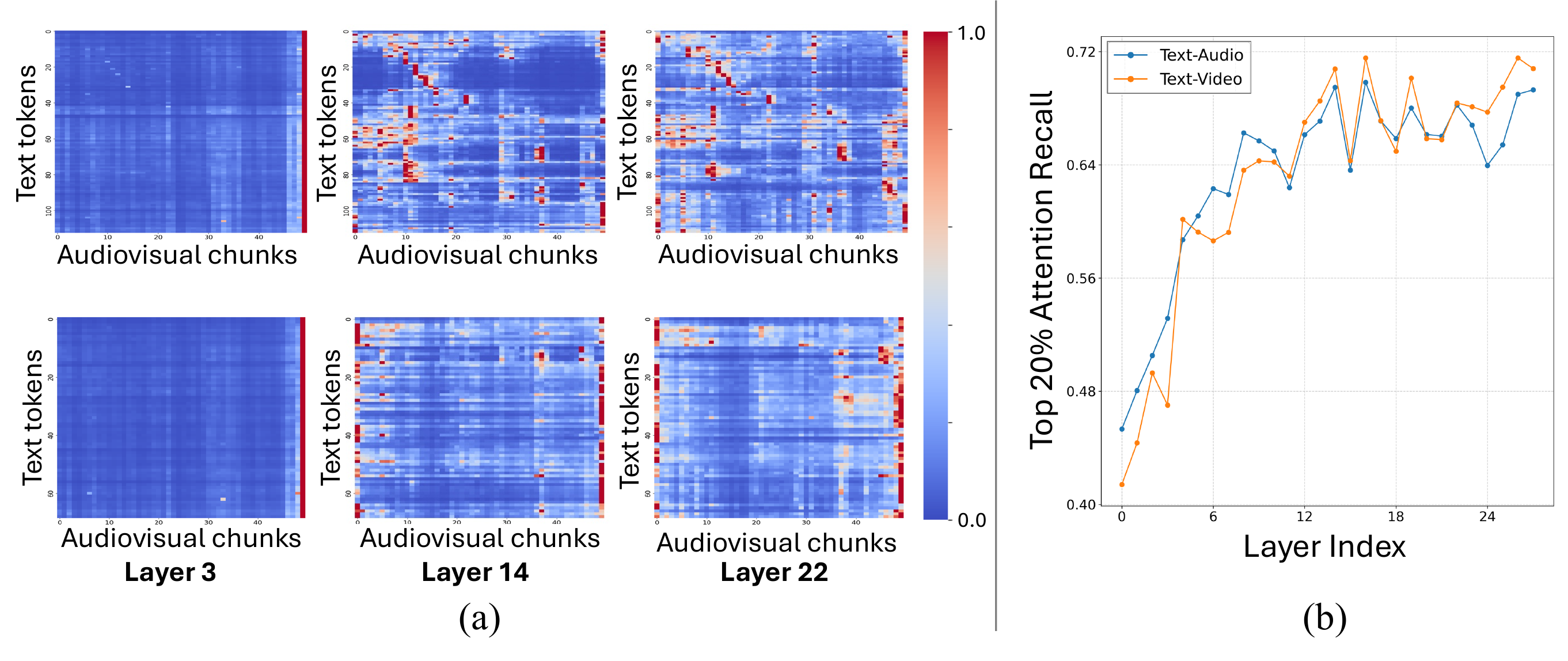}
  \caption{(a) Layer-wise text-to-audiovisual attention scores, averaged within each audiovisual chunk and normalized for better visualization. We show representative decoder layers, i.e., Layers 3, 14, and 22. The first and second rows correspond to the queries ``\textit{Who said in the video that `...'?}'' and ``\textit{In which part of the video does the man with ... appear?}'', respectively.
(b) Top-20\% attention recall
of text-to-audio and text-to-video attention scores across decoder layers.}
  \label{fig:layerwise}
\end{figure}

As shown in Fig.~\ref{fig:layerwise} (a), the model gradually shifts its attention toward the audiovisual chunks corresponding to the correct answer for both queries. For example, in the top row, the correct answer corresponds to a person appearing in the early part of the video, and the model progressively concentrates its attention on the corresponding early chunks. In contrast, in the bottom row, the answer is grounded in a segment appearing later in the video, and the attention accordingly shifts toward the later chunks. This indicates that the model naturally identifies query-specific evidence through decoder-layer processing, highlighting that the importance of audiovisual token is \textit{query-dependent}.

\textbf{Attention sparsity is layer-dependent.}
We define top-20\% attention recall~\cite{minference,fastkv} as the fraction of total attention mass captured by the top 20\% largest attention values, where system prompts are excluded to focus on semantically meaningful interactions. 
Text-to-audio and text-to-video attention maps are averaged over 64 sample videos from the WorldSense benchmark~\cite{worldsense}. 

As shown in Fig.~\ref{fig:layerwise} (b), the top-20\% attention recall for both modalities remains below 0.6 in the early decoder layers.
At this stage, text tokens actively attend to a broad range of tokens regardless of modality.
Through this process, audiovisual evidence is aggregated into the updated text representations.
As this process repeats across decoder layers, the text representations progressively contains integrated audio-video information.
Consequently, it remains difficult to identify a small set of clearly text-relevant tokens in the early layers.
As the representations pass through the decoder, the attention recall gradually increases, reaches its first peak in the middle layers and remains relatively high in the later layers. This trend suggests that text-to-audiovisual attention becomes progressively more selective after the middle layers, allowing the model to better distinguish informative tokens from less relevant ones.

These observations motivate us to propose a layer-adaptive pruning strategy. Before the middle layers, token pruning should be performed conservatively to allow sufficient omni-modal information integration. Once the model begins to exhibit clearer text-relevant token preferences from the middle layers, pruning can be applied more aggressively, targeting tokens that receive concentrated attention from text.

\section{OmniDrop}

\begin{figure}[t]
  \centering
  \includegraphics[width=0.9\textwidth]{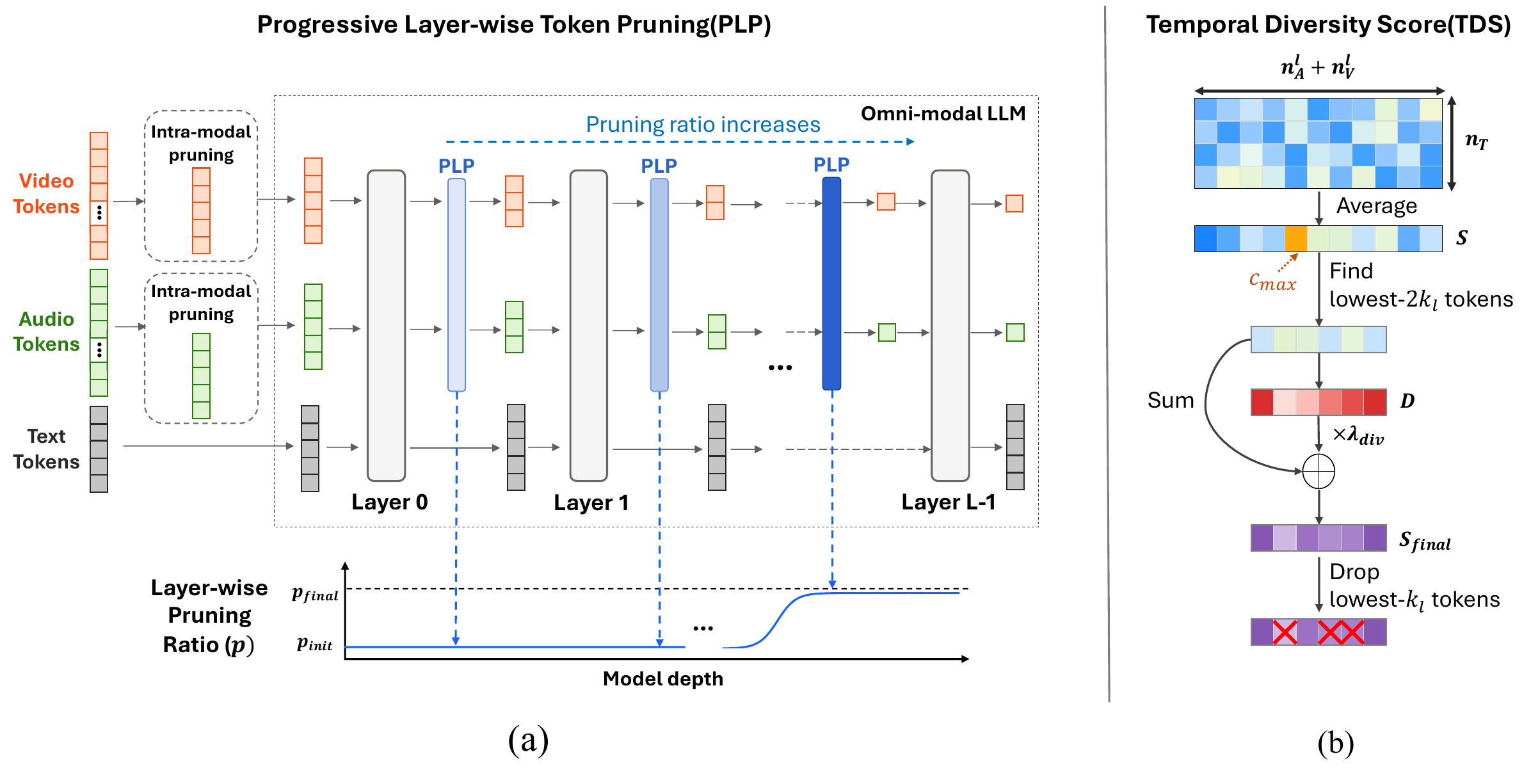}
  \caption{Overview of \mymethod. (a) Progressive layer-wise token pruning (PLP) schedule based on the sigmoid function (\cref{eq:prune_ratio,eq:sigmoid_schedule}). (b) Query-guided token importance (Eq. \ref{eq:importance}) with a temporal diversity score (Alg. \ref{alg:tds}) for audiovisual tokens at each layer.}
  \label{fig:main}
\end{figure}

We propose \mymethod~that enhances the efficiency of Omni-modal LLMs via adaptive layer-wise token pruning. Unlike embedding-level compression methods, \mymethod~dynamically adjusts audio-video token retention by leveraging text-based semantic guidance and temporal diversity.

\subsection{Progressive layer-wise token pruning}  
\label{plp}
As analyzed in Sec. \ref{sec:revisit}, audio and video embeddings are not sufficiently aligned before entering the LLM, but become progressively integrated throughout the decoder layers. 
Since the model has not yet formed reliable query-relevant token preferences in the early-to-middle layers, a pruning strategy should preserve more tokens in shallow layers and apply stronger reduction in deep layers once query-relevant evidence becomes more clearly identifiable.

We thus propose a \textbf{progressive layer-wise token pruning (PLP)} strategy. To reflect that informative audiovisual tokens become progressively more distinguishable across decoder layers, we define the dynamic pruning ratio $p_l$ for the $l$-th layer as follows: 
\begin{equation}
\label{eq:prune_ratio}
p_l = p_{init} + (p_{final} - p_{init}) \cdot \sigma(l, t_{mid}, \beta),
\end{equation}

where $p_{init}$ and $p_{final}$ denote the initial and final pruning ratios, respectively.
The transition between these ratios is scheduled by a sigmoid function applied over the $L$ total layers, defined as 
\begin{equation}
\sigma(l, t_{mid}, \beta) = \frac{1}{1 + \exp(-\beta \cdot (\frac{l}{L-2} - t_{mid}))},
\label{eq:sigmoid_schedule}
\end{equation}
where $\beta$ and $t_{mid}$ represent the slope parameter and  the midpoint of the sigmoid function, respectively. As shown in \cref{fig:main} (a), PLP is applied up to the penultimate layer (\textit{i.e.}, Layer $L-2$) to preserve the hidden states immediately preceding the language modeling head.
This schedule, based on our observations in Sec. \ref{sec:analysis}, is designed to preserve sufficient information in early layers for omni-modal fusion, while aggressively pruning redundant tokens from the middle layers onward.

\subsection{Query-guided token importance}
To eliminate less informative audio and video tokens, we utilize text queries as guidance. 
By leveraging text information, we can selectively retain only the \textit{task-relevant} tokens required by the query. 
Within a fixed token budget, the model adaptively allocates more tokens to the modality that is more relevant to the task—retaining more video tokens for video-centric tasks and more audio tokens for audio-centric tasks. For tasks that require both modalities, the model preserves a balanced distribution between them.

The query-guided token importance score $S$ is calculated as follows:
\begin{equation}
\label{eq:importance}
  S = \frac{1}{n_T} \sum_q^{n_T} \mathrm{Attn}_{q \rightarrow \{V, A\}},
\end{equation}
where $n_T$ is the number of text tokens and $\mathrm{Attn}_{q \rightarrow \{V, A\}}$ represents the attention score map from the current text token $q$ to the video ($V$) and audio ($A$) tokens. The score \(S\) indicates how strongly the text tokens, on average, attend to audio and video tokens.
To reduce the length of the sequence, we remove the $k_l$ tokens with the lowest scores $S$ according to the pruning ratio $p_l$ defined in~\cref{eq:prune_ratio}.
For each layer $l$, the number of pruned tokens $k_l$ is determined by
\begin{equation}
  k_l = \left\lfloor (n_A^l + n_V^l) \cdot p_l  \right\rfloor,
\end{equation}
where $n_A^l$ and $n_V^l$ represent the total number of audio and video tokens present at the $l$-th layer, respectively, and $\left\lfloor\cdot\right\rfloor$ denotes the floor function. 
The pruning intensity follows the predefined sigmoid schedule, targeting the least semantically relevant tokens.

\subsection{Temporal diversity-aware token pruning}

\begin{algorithm}[t]
\caption{Temporal Diversity Score (TDS)}
\label{alg:tds}
\begin{algorithmic}[1]
\REQUIRE Importance scores $S$, chunk indices $C$, pruning budget $k_l$, diversity weight $\lambda_{div}$
\ENSURE Set of tokens to prune $\mathcal{P}$

\STATE $c_{max} \leftarrow \text{chunk\_idx}(\text{argmax } S)$ \COMMENT{Identify the key chunk}
\STATE $\mathcal{I}_{cand} \leftarrow \text{indices of }2k_l \text{-lowest scores in } S$ \COMMENT{Candidate buffer}

\STATE $\{D_i\}_{i=1}^{2k_l} \leftarrow |c_{max} - c_i| / \max(C)$ \COMMENT{Normalized temporal distance}
\STATE $\{S_{TDS, i}\}_{i=1}^{2k_l} \leftarrow S_i + \lambda_{div} \cdot D_i$ \COMMENT{Score augmentation}

\STATE $\mathcal{P} \leftarrow \text{indices of } k_l\text{-lowest tokens based on } S_{TDS}$
\RETURN $\mathcal{P}$
\end{algorithmic}
\end{algorithm}

While text-based importance scoring ($S$) effectively identifies query-relevant tokens, an over-reliance on text attention can lead to capturing only fragmentary information. In tasks that require an understanding of long-form temporal dynamics or the sequential context of events, focusing solely on high-text-attention segments may result in a loss of overall context. To address this, we introduce a \textit{temporal diversity score} (TDS), designed to ensure a balanced representation of the entire sequence while pruning.
The procedure is summarized in Alg. \ref{alg:tds} and consists of the following steps:
\begin{enumerate}[leftmargin=*, label=\arabic*.]
    \item \textbf{Identify the key chunk ($c_{max}$) (line 1):} 
    We first determine the chunk index $c_{max}$ that includes the token achieving the highest importance score $S$ in the current layer.
    \item \textbf{Candidate selection (line 2):} Rather than pruning $k_l$ tokens immediately, we select an expanded candidate set $\mathcal{I}_{cand}$ consisting of the $2k_l$ tokens with the lowest scores based on $S$.
    \item \textbf{Distance-based diversity scoring (line 3):} 
    For all tokens in the candidate set, we compute a normalized distance factor $D$ relative to the key chunk $c_{max}$, defined as $D = \frac{|c_{max} - c_i|}{\max(C)}$, where $c_i$ represents the chunk index of token $i$, and $\max(C)$ denotes the maximum chunk index in the sequence. 
    
    \item \textbf{Final score integration (line 4):} The final importance score $S_{TDS}$ is defined as the weighted sum of the initial score $S$ and the diversity factor $D$: $S_{TDS} = S + \lambda_{div} \cdot D$, where $\lambda_{div}$ is a hyperparameter controlling the weight of diversity. Thus, tokens located temporally farther from the primary attention peak ($c_{max}$) receive a score boost, making them less likely to be pruned. 
\end{enumerate}

Based on $S_{TDS}$, we drop the $k_l$-lowest tokens. 
Consequently, our method prioritizes dropping tokens that are both textually irrelevant and temporally redundant relative to the most significant event.

\subsection{Intra-modality token pruning prior to LLM}
\label{sec:intra}
Previous studies~\cite{omnizip, dycoke, speechprune} have shown that each modality often contains tokens that are either non-salient or redundant, which unnecessarily increase the computational burden of Omni-LLMs. In addition, as shown in Fig.~\ref{fig:cosine_sim} (b), some intra-modal token pairs exhibit high cosine similarity, indicating substantial redundancy within each modality.
Therefore, before feeding audiovisual tokens into the LLM, we first perform independent modality-specific pruning to improve inference efficiency while preserving informative content.
For audio tokens, we follow OmniZip~\cite{omnizip}, which retains only tokens with the highest attention scores from the last layer of the audio encoder, assuming that highly attended tokens carry the most salient acoustic information.
For video tokens, we adopt the Temporal Token Merging (TTM) strategy used in Dycoke~\cite{dycoke}, which filters out consecutive and redundant visual tokens based on cosine similarity between them. This step effectively compresses repeated visual content while maintaining essential scene dynamics. After intra-modality pruning, 70\% of audio tokens and 40\% of video tokens remain, resulting in a total retention of 45\% of tokens.

\section{Experiments}
To evaluate the performance of our framework, we conduct experiments on several representative benchmarks: VideoMME~\cite{videomme} for video understanding, WorldSense~\cite{worldsense} for question-answering requiring audiovisual integration, and AVUT~\cite{avut} for audio-centric video understanding.
% To evaluate the performance of our framework, we conduct experiments on several representative benchmarks: VideoMME~\cite{videomme} for video understanding, WorldSense~\cite{worldsense} for question-answering requiring audiovisual integration, AVUT~\cite{avut} for audio-centric video understanding, and ShortVid-Bench~\cite{shortvid} for short-form temporal reasoning.
We evaluate our proposed framework on the Qwen2.5-Omni~\cite{qwen} (7B and 3B) models, publicly available Omni-LLMs, comparing with existing SOTA training-free methods: OmniZip~\cite{omnizip} and DASH~\cite{dash}. 

Since \mymethod~employs progressive layer-wise pruning, we report the average token retention ratio across all layers to ensure a fair comparison with fixed-ratio baseline methods.
We configure the pruning schedule ($p_{init}, p_{final}$) based on the target retention: ($0.0, 0.2$) and ($0.02, 0.5$) for 30\% and 20\% on the 7B model, and ($0.0, 0.15$) and ($0.02, 0.5$) for the 3B variant, respectively.
We set the sigmoid parameters at $t_{mid}=0.5$ and $\beta=20$, except for the 3B model's 20\% retention configuration, where $t_{mid}$ is $0.55$. The weight for the diversity score, $\lambda_{div}$, is set to 0.2. The diversity-aware mechanism is applied from the mid-section of the decoder layers (layer 14 for the 7B model and layer 19 for the 3B model).

We conduct all experiments on a single NVIDIA H100 GPU and employ FlashAttention~\cite{flashattn} to reduce memory usage.
We provide the hyperparameters of other baselines and more experimental details in the Appendix~\ref{sec:appendixA}.

\begin{table}[t]
\caption{Comprehensive performance and efficiency comparison of different methods on Qwen2.5-Omni models. Avg. $\Delta_{\text{Perf.}}$ denotes the average performance change across four benchmarks when reducing the retained ratio from 30\% to 20\%.}
\centering
\label{tab:combined_performance}
\resizebox{\textwidth}{!}{
\begin{tabular}{llcccccccc}
\toprule
\textbf{Method} 
& \makecell{\textbf{Retained} \\ \textbf{Ratio (\%)}} 
& \textbf{VideoMME} 
& \textbf{WorldSense} 
& \textbf{AVUT} 
& \makecell{\textbf{Avg.} \\ $\Delta_{\text{Perf.}}$}
& \makecell{\textbf{Prefill} \\ \textbf{Time (s)}} 
& \makecell{\textbf{Latency /} \\ \textbf{Sample (s)}} 
& \makecell{\textbf{GPU Mem} \\ \textbf{(GB)}} \\
\midrule

% ==================== 7B Section ====================
\textit{Qwen2.5-Omni-7B} & 100 & 64.67 & 46.85 & 65.17 & -- & 1.73 & 1.82 & 28.92 \\
\midrule

\multirow{2}{*}{OmniZip~\cite{omnizip}} 
& 30 & 65.85 & 45.55 & 61.76 & -- & 1.06 & 1.15 & 25.76 \\
& 20 & 64.85 & 44.10 & 61.07 & -1.08 & \textbf{0.97} & \textbf{1.05} & 25.65 \\
\midrule

\multirow{2}{*}{DASH~\cite{dash}} 
& 30 & 65.67 & 45.87 & 60.96 & -- & 1.07 & 1.15 & 25.68 \\
& 20 & 64.96 & 44.61 & 60.09 & -1.09 & 1.02 & 1.11 & 25.65 \\
\midrule

\multirow{2}{*}{\makecell[l]{\mymethod}}
& 30 & \textbf{66.52} & \textbf{46.60} & \textbf{64.01} & -- & 1.05 & 1.15 & \textbf{25.65} \\
& 20 & 66.44 & 46.50 & 63.67 & \textbf{-0.18} & 1.04 & 1.14 & \textbf{25.65} \\
\midrule[1.5pt]
% ==================== 3B Section ====================
\textit{Qwen2.5-Omni-3B} & 100 & 62.51 & 46.63 & 62.40 & -- & 1.35 & 1.40 & 18.63 \\
\midrule

\multirow{2}{*}{OmniZip~\cite{omnizip}} 
& 30 & 62.52 & 45.71 & 60.27 & -- & 0.95 & 0.99 & 15.98 \\
& 20 & 61.59 & 44.99 & 59.28 & -0.98 & \textbf{0.94} & \textbf{0.98} & 15.91 \\
\midrule

\multirow{2}{*}{DASH~\cite{dash}} 
& 30 & 62.26 & 44.83 & 60.21 & -- & 1.03 & 1.08 & 15.91 \\
& 20 & 60.96 & 44.10 & 58.48 & -1.36 & 0.97 & 1.02 & 15.91 \\
\midrule

\multirow{2}{*}{\makecell[l]{\mymethod}}
& 30 & \textbf{63.07} & \textbf{46.78} & \textbf{61.53} & -- & 0.99 & 1.03 & \textbf{15.90} \\
& 20 & \textbf{63.07} & 46.19 & 60.55 & \textbf{-0.47} & 0.97 & 1.02 & \textbf{15.90} \\

\bottomrule
\end{tabular}
}
\vspace{0.1cm}
\end{table}

\subsection{Comparison with baselines}
\label{sec:baseline}

Tab. \ref{tab:combined_performance} compares \mymethod~with OmniZip and DASH across multiple benchmarks. 
We report the average accuracy of all domains for both VideoMME and AVUT, with detailed scores provided in the Appendix \ref{sec_a:deatail}.
Overall, \mymethod~consistently achieves superior performance under aggressive compression while maintaining high efficiency. 
Notably, at 30\% retention, \mymethod~outperforms all baselines and even surpasses the 100\% full-token baseline on VideoMME (66.52 vs. 64.67 for 7B; 63.07 vs. 62.51 for 3B). This advantage persists even at an extreme 20\% retention ratio, with \mymethod~consistently outperforming existing baselines and achieving a particularly large gain of 3.58 points on AVUT under the 7B setting (63.67 vs. 60.09).

The robustness of \mymethod~is further highlighted by its minimal performance degradation when scaling down from 30\% to 20\% retention. The average drop for \mymethod~is only 0.18 (7B) and 0.47 (3B) points, significantly lower than the drops exceeding 1.0 point observed in OmniZip and DASH. These results, consistent across both 7B and lightweight 3B models, demonstrate that our framework effectively preserves critical omni-modal information even under extreme token reduction.

\mymethod~also delivers substantial gains in inference efficiency on the WorldSense benchmark.
It reduces prefill time by 39.9\% (7B) and 28.1\% (3B), while decreasing GPU memory footprint by 11.3\% and 14.7\%, respectively. 
These results confirm that \mymethod~improves the memory-efficiency of Omni-LLM inference without compromising task accuracy. 
Although \mymethod~retains approximately 45\% of tokens in the early layers, its highly aggressive pruning in the later layers keeps the average GPU memory usage and latency nearly comparable to those of other baselines that remove a large fraction of tokens at the input-level.

\subsection{Ablation study}

\begin{wraptable}{r}{0.48\linewidth}
    \vspace{-15pt}
    \caption{Ablation studies on the PLP schedule and TDS.}
    \vspace{0.1cm}
    \centering
    \label{tab:ablation}
    \small
    \setlength{\tabcolsep}{4pt}
    \begin{tabular}{lccc}
    \toprule
    Method & Ratio & WS & AVUT \\
    \midrule
    Intra-pruning & 45 & 46.50 & 64.19 \\
    Intra-pruning & 30 & 44.33 & 59.34 \\
    \midrule
    PLP - Exp & 30 & 46.53 & 63.90 \\
    PLP - Sig & 30 & 46.53 & 63.84 \\
    + TDS (Ours) & 30 & \textbf{46.60} & \textbf{64.01} \\
    \midrule
    PLP - Exp & 20 & 45.93 & 63.03 \\
    PLP - Sig & 20 & 46.25 & 63.32 \\
    + TDS (Ours) & 20 & \textbf{46.50} & \textbf{63.67} \\
    \bottomrule
    \end{tabular}

    \vspace{15pt}

    \caption{Ablation studies on guidance type.}
    \vspace{0.1cm}
    \centering
    \label{tab:a2v}
    \small
    \setlength{\tabcolsep}{4pt}
    \begin{tabular}{lccc}
    \toprule
    Method & Ratio & WS & AVUT \\
    \midrule
    Audio guidance & 30 & 45.59 & 60.84 \\
    Text guidance (Ours) & 30 & \textbf{46.78} & \textbf{61.53} \\
    \midrule
    Audio guidance & 20 & 43.25 & 56.57 \\
    Text guidance (Ours) & 20 & \textbf{46.19} & \textbf{60.55} \\
    \bottomrule
    \end{tabular}

    \vspace{-10pt}
\end{wraptable}

\textbf{Progressive layer-wise token pruning (PLP).}
Tab. \ref{tab:ablation} presents ablation studies on Qwen2.5-Omni-7B to evaluate the effectiveness of each component in \mymethod. 
Compared with the baseline that applies only intra-modality token pruning, PLP substantially improves performance under tighter token budgets. At the 30\% retention ratio, the baseline drops to 44.33 on WorldSense and 59.34 on AVUT, whereas PLP restores performance to 46.53 and 63.90, respectively. This demonstrates that layer-wise pruning is significantly more effective than applying compression before LLM injection.
When PLP is additionally applied using either exponential 
($p_l = p_{init} \times (p_{final}/p_{init})^{\frac{l}{N-2}}$) 
or sigmoid decay at the 30\% ratio, both schedules yield nearly identical performance. However, the difference is more pronounced under the more aggressive 20\% setting. In this case, the sigmoid schedule consistently outperforms the exponential schedule, improving WorldSense from 45.93 to 46.25 and AVUT from 63.03 to 63.32. These results indicate that a schedule that preserves most tokens in the early layers and performs more aggressive pruning from the middle layers onward is more effective for maintaining performance under extreme compression.

\textbf{Temporal diversity score (TDS).}
In Tab. \ref{tab:ablation}, TDS further improves results at all token budgets. At the 30\% ratio, TDS increases WorldSense from 46.53 to 46.60 and AVUT from 63.84 to 64.01. Under the more aggressive 20\% ratio, the gains become larger, improving WorldSense from 46.25 to 46.50 and AVUT from 63.32 to 63.67. These results indicate that preserving temporally diverse tokens becomes increasingly important as the compression ratio becomes more aggressive.

\textbf{Query-guided token importance.}
While existing baselines~\cite{omnizip, dash} compress video tokens based on audio guidance, we further evaluate whether audio-to-video attention can serve as a reliable criterion for token importance. Specifically, we replace text-to-audiovisual attention with audio-to-video attention when computing the token importance score $S$. However, constructing the audio-to-video attention map incurs substantial computational overhead, causing Out-of-Memory errors on Qwen2.5-Omni-7B. We therefore perform this ablation using the 3B model. As shown in Tab.~\ref{tab:a2v}, audio-guided pruning consistently degrades performance, with a particularly large drop under the 20\% retained-token setting. This suggests that text-guided attention provides a more reliable signal for distinguishing important audiovisual tokens.

\begin{figure}
  \centering
  \includegraphics[width=\textwidth]{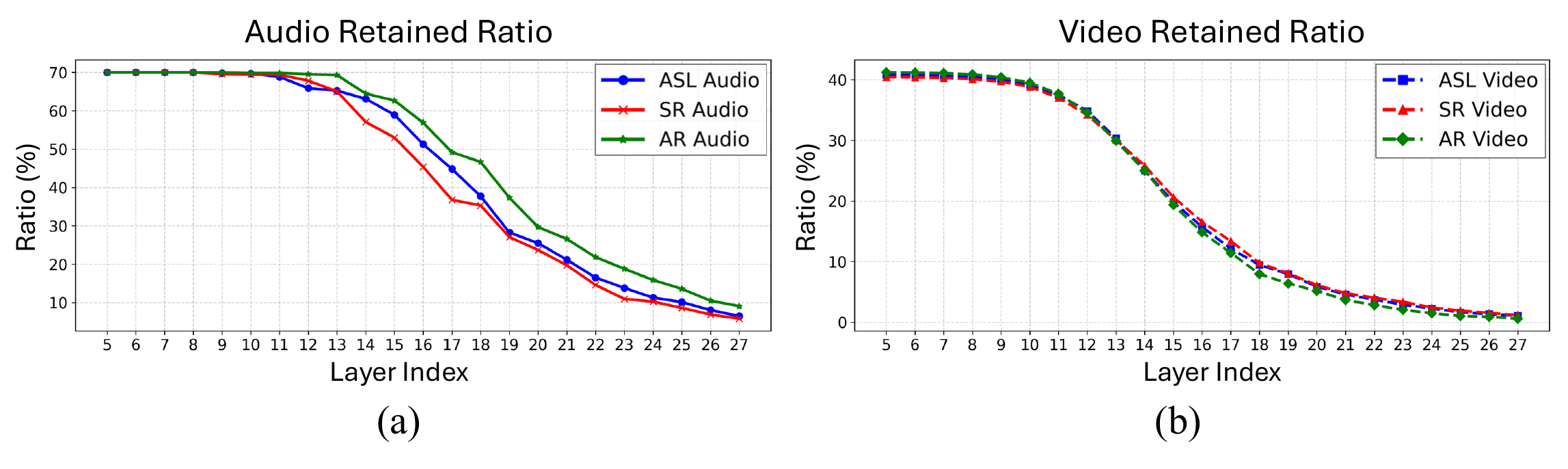}
  \caption{Comparison of audio/video retained ratio per decoder-layer across three different tasks,\textit{ Audio Source Localization} (ASL), \textit{Scene Recognition} (SR), and \textit{Audio Recognition} (AR).  (a) Audio retained ratios across tasks. (b) Video retained ratio across tasks. Our query-guided pruning enables retention ratios adapted to task requirements without explicit task labels.}
  \label{fig:av-ratio}
  \vspace{-12pt}
\end{figure}

\subsection{Task-adaptive token retention via query-guidance}
To verify whether our query-guided token importance facilitates task-adaptive pruning, we analyze the token retention behavior across three distinct tasks from the WorldSense benchmark: \textit{Scene Recognition} (SR), \textit{Audio Recognition} (AR), and \textit{Audio Source Localization} (ASL). These tasks represent varying degrees of modality-specific information: SR is a visual-centric task where the model must identify environments based on visual cues. AR is an audio-centric task focused on identifying sounds or speech. ASL requires the integration of both auditory and visual signals to pinpoint a sound source.

Fig. \ref{fig:av-ratio} illustrates the retained ratio of audio and video tokens across LLM decoder layers for each task. In Fig. \ref{fig:av-ratio}(a), which displays audio token retention, we observe that the AR task maintains the highest ratio, followed by ASL and SR. This confirms that for audio-centric tasks, our scoring mechanism prioritizes the preservation of auditory information. Conversely, (b) shows that the video token retention ratio is highest for SR, followed by ASL, and lowest for AR. Notably, the ASL task, which necessitates both modalities, maintains a balanced retention of tokens compared to the modality-specific tasks. These results demonstrate that our query-guided scoring effectively ensures task-adaptive token retention.

\section{Conclusion}

We have proposed~\mymethod, a training-free, layer-wise token pruning framework for efficient Omni-LLMs. By using query texts as guidance, \mymethod~enables task-relevant and modality-adaptive token retention. Our progressive layer-wise pruning preserves early omni-modal integration while removing non-salient tokens in deeper layers, and a temporal diversity score prevents the loss of global context caused by focusing solely on highly query-relevant tokens. Extensive experiments on multiple audiovisual benchmarks demonstrate that \mymethod~consistently outperforms existing token compression methods by up to 3.58 points while reducing prefill latency by up to 40\% and memory consumption by up to 14.7\%.

\textbf{Limitations and future work.}
Despite these encouraging results, several limitations remain. First, \mymethod~currently depends on text prompts for token selection, which restricts its applicability in scenarios where only audio and video inputs are available. Developing methods that directly capture semantic relationships between audio and video without textual guidance remains an important open challenge. Second, as a training-free framework, several components of \mymethod, including pruning schedules and hyperparameter choices, are determined empirically. Future work may explore whether these decisions can be learned or jointly optimized using calibration data to further improve robustness and generalization.

\clearpage

\bibliography{references}
\bibliographystyle{plain}
\medskip

\small

%%%%%%%%%%%%%%%%%%%%%%%%%%%%%%%%%%%%%%%%%%%%%%%%%%%%%%%%%%%%%%
\newpage
\appendix

\section*{Appendices}

\section{Additional experimental details}
\label{sec:appendixA}
We follow the preprocessing settings of OmniZip~\cite{omnizip} and DASH~\cite{dash}. To handle the longer videos in VideoMME~\cite{videomme} and retain finer temporal resolution, we set the maximum number of input frames to 768. For all other datasets, we use a maximum of 128 input frames. Each audiovisual chunk contains 50 audio tokens and 288 video tokens.

For intra-modality pruning, we use different hyperparameters for each modality. We set the audio pruning ratio to 0.3. For video token pruning, we apply a temporal token merging (TTM) strategy to every four consecutive frames with a pruning rate of 0.8, resulting in an overall video retention ratio of 40\%.

We utilize the official implementations of the existing methods for comparative analysis.
In our experiment with OmniZip\footnote{OmniZip: \url{https://github.com/KD-TAO/OmniZip}}, we set $\rho_a=0.45$ and $\rho_v=0.75$ for the 30\% retention, and $\rho_a=0.45$ and $\rho_v=0.85$ for the 20\%. Also, we adjust the upper and lower limits of the pruning rate ($\rho_{max}$, $\rho_{min}$) to prevent excessive pruning and align with our target retention ratios: (0.85, 0.40) for the 30\% target and (0.95, 0.45) for the 20\%.
Other hyperparameters were kept unchanged.
With DASH\footnote{DASH: \url{https://github.com/laychou666/DASH}}, we maintain original configuration, except that the target retention ratios $\rho_a$ and $\rho_v$ are set to the same values as those used for OmniZip.

For visualization in Fig.~\ref{fig:cosine_sim}, we sample 64 videos from WorldSense~\cite{worldsense} and extract audio and video embeddings using the corresponding encoders.
For the PCA visualization in Fig.~\ref{fig:cosine_sim} (a), we first fit PCA with two components using a balanced subset of audio and video tokens.
We then transform all tokens using the fitted PCA.
For visualization, we sample 10,000 tokens according to the original modality ratio, resulting in 1,392 audio tokens and 8,608 video tokens.
For the cosine similarity analysis in Fig.~\ref{fig:cosine_sim} (b), we compute all pairwise cosine similarities within each video for audio-audio, video-video, and audio-video token pairs.
The resulting similarities are flattened and concatenated across all 64 videos.
For KDE visualization, we randomly sample 100,000 values from each distribution.

\section{Related works and comparison with OmniZip and DASH}
\label{sec:appendixB}

Understanding modality-specific redundancy inherent in the input data is crucial to improving the efficiency of multimodal LLMs. For example, He et al.~\cite{mae} shows that a significant portion of visual patches can be omitted without losing global context. Similarly, audio signals exhibit temporal and spectral redundancy, often reflected in strong local correlations in spectrograms~\cite{audio-mae}. This gap between raw data representation and actual information density provides strong motivation for pruning or merging redundant tokens.

Token compression methods have also been widely studied by exploiting modality-specific redundancy. In the visual domain, spatial redundancy among image patches is reduced by pruning low-importance tokens~\cite{fastv}, merging similar ones~\cite{ToMe}, or selecting informative subsets~\cite{DivPrune}. These concepts extend to video through dynamic frame or patch removal~\cite{dycoke} and to audio via attention-guided pruning~\cite{speechprune}. 
However, uni-modal strategies are insufficient for Omni-LLMs, where independent compression poses a potential risk of disrupting temporal correspondence and cross-modal alignment. This necessitates token selection based on cross-modal interaction.
Therefore, recent Omni-LLM token compression  have begun to incorporate cross-modal cues when determining token importance or redundancy.

Recent Omni-LLM token compression methods~\cite{omnizip,omnisift,echoing,dash} have explored different ways to reduce the computational cost of processing long audio-video inputs. Some approaches, such as OmniSift~\cite{omnisift} and EchoingPixels~\cite{echoing}, rely on additional training to learn token selection or allocation policies. In contrast, another line of work~\cite{omnizip, dash} aims to compress tokens in a training-free manner, making them easier to apply to existing Omni-LLMs without modifying model parameters.

Below, we focus on training-free Omni-LLM token compression methods, OmniZip~\cite{omnizip} and DASH~\cite{dash}, and provide a detailed comparison with our approach.

\textbf{OmniZip}~\cite{omnizip} is an audio-guided audiovisual token compression framework. It first uses attention scores from the audio encoder to identify important audio tokens. Less important audio tokens are not simply discarded; if they are highly similar to video tokens, they are merged into nearby anchor tokens. For video tokens, OmniZip adopts Interleaved Spatio-Temporal Compression, which alternately reduces temporal redundancy by merging cross-frame tokens and spatial redundancy by pruning intra-frame tokens. In addition, the retention ratio of video tokens is determined based on the saliency scores of the retained audio tokens, allowing video pruning to be guided by audio importance. Therefore, OmniZip relies on the assumption that audio and video tokens with high input-level similarity are semantically related or redundant.

\textbf{DASH}~\cite{dash} compresses audiovisual tokens by exploiting semantic structure. It first detects semantic boundaries in the audio stream using intra-audio cosine similarity. These audio-derived boundaries are then projected onto the video stream to perform dynamic segmentation. This step assumes that audio and video signals occurring at the same time tend to share the same semantic context (\textit{e.g.} scene changes, topic transitions). DASH then selects important audio tokens using tri-signal fusion, which combines boundary probability, density, and attention score. For video tokens, DASH performs boundary-aware pruning, where tokens near likely semantic boundaries are assigned higher importance and are more likely to survive. Similar to OmniZip, DASH uses the audio retention rate as a proxy for segment-level information density to determine the video token retention ratio.

However, both methods rely on audio-derived signals as a proxy for cross-modal token importance, which can fail under weak audio-video correlation, noise, or task-irrelevant co-occurrences.

\textbf{OmniDrop} does not rely on input-level audio-video similarity or temporal co-occurrence to determine modality correlation. Instead, OmniDrop uses the text query as a task-specific pruning signal. Specifically, audio and video tokens are evaluated according to their attention with the text query, and tokens that show low attention to the query are considered less relevant and are pruned regardless of their modality. This allows OmniDrop to remove unnecessary audio and video tokens based on their contribution to the target query, rather than based on predefined audio-video similarity assumptions.

OmniDrop further introduces a temporal diversity score (TDS). Since tokens around key chunks, which strongly interact with the text query, are already likely to survive, OmniDrop gives relatively higher scores to tokens that are temporally farther from such key chunks. 

Beyond the scoring mechanism, OmniDrop fundamentally differs from prior work in \textit{where} pruning occurs. While OmniZip and DASH operate at the input embedding level, OmniDrop performs pruning progressively within the LLM decoder layers. This design is enabled by our text-guided scoring, which remains valid even after omni-modal fusion. In contrast, the scoring mechanisms of OmniZip and DASH are inherently tied to the input level. OmniZip's audio-video cosine similarity assumes that the two modalities remain directly comparable, and DASH assumes audio-derived semantic boundaries. However, once self-attention layers progressively fuse audio and video representations, both assumptions break down: cosine similarity no longer reflects cross-modal relationships, and intra-audio similarity no longer provides reliable boundary. Thus, OmniDrop's text-guided scoring is not merely an alternative criterion but \textit{a necessary design choice for unlocking layer-wise pruning in Omni-LLMs}.

\begin{table*}[t]
\centering
\caption{Comparison of OmniZip, DASH, and OmniDrop.}
\label{tab:method_comparison}
\renewcommand{\arraystretch}{1.3}
\small
\begin{tabularx}{\textwidth}{l XXX}
\toprule
\textbf{Feature} & OmniZip~\cite{omnizip} & DASH~\cite{dash} & \textbf{OmniDrop [Ours]} \\
\midrule
\textbf{Main Idea} & Audio-guided  & Structure-aware & Text-guided layer-wise \\
\midrule
\textbf{Pruning Stage} & Input-level & Input-level & Progressive Layer-wise \\
\midrule
\textbf{Key Metrics} & 1.Audio self-attn. & 1.Boundary probability & 1.Query to (A/V) attn.\\ 
     & 2.A-V Cos similarity & 2.Density-based uniqueness & 2.Temporal Distance Score \\
     &  & 3.Audio self-attn. &  \\
\midrule
\textbf{Assumption} & A-V similarity denotes semantic relevance & Co-occurrence shares context & Text query defines relevance; loosely coupled at input-level \\
\bottomrule
\end{tabularx}
\end{table*}

\section{Detailed performance of WorldSense and AVUT}
\label{sec_a:deatail}
Due to page limitations, Sec.~\ref{sec:baseline} reports only the average performance on the WorldSense~\cite{worldsense} and AVUT~\cite{avut} benchmarks. In this section, we provide detailed scores for each domain and sub-task for a more fine-grained analysis.

\begin{table}[ht]
\centering
\caption{Performance comparison of various methods on the WorldSense~\cite{worldsense} benchmark.}
\label{tab:supp_ws_app}
\resizebox{\textwidth}{!}{
\small
\begin{tabular}{lcccccccccc}
\toprule
Method & Ret. Ratio (\%) & Music & Cult. & Tech & Daily & Film & Sports & Perf. & Games & Avg. \\
       &            &       & \& Pol. & \& Sci. & Life & \& TV &        &       &       &      \\
\midrule
\textit{Qwen2.5-Omni-7B} & 100 & 48.77 & 53.40 & 50.61 & 46.81 & 44.06 & 42.70 & 43.07 & 43.35 & 46.85 \\
\midrule
OmniZip~\cite{omnizip} & 30 & 46.06 & 52.43 & 49.59 &\textbf{ 47.42} & 41.69 & 41.16 & \textbf{43.45} & 38.63 & 45.55 \\
 & 20 & 45.57 & 49.84 & 48.37 & 46.05 & 40.37 & 38.84 & 41.20 & 38.63 & 44.10 \\
\midrule
DASH~\cite{dash} & 30 & 46.06 & 51.13 & \textbf{51.84} & 46.20 & \textbf{43.27} & 43.26 & 41.57 & 39.06 & 45.87 \\
 & 20 & 47.29 & 50.81 & 48.98 & 45.14 & 40.63 & 41.86 & 38.58 & 39.48 & 44.61 \\
\midrule
OmniDrop & 30 & \textbf{47.54} & 53.07 & 50.82 & 46.81 & 43.01 & \textbf{43.72} & 41.95 & 43.35 & \textbf{46.60} \\
 & 20 & 46.80 & \textbf{53.40} & 50.61 & 46.50 & 42.74 & \textbf{43.72} & 42.70 & \textbf{43.78} & 46.50 \\
\midrule[1.5pt]
\textit{Qwen2.5-Omni-3B} & 100 & 46.06 & 51.46 & 51.02 & 45.74 & 45.38 & 44.88 & 45.69 & 40.77 & 46.63 \\
\midrule
OmniZip~\cite{omnizip} & 30 & 43.10 & 50.49 & 51.02 & \textbf{47.11} & 44.06 & 41.63 & 42.70 & 42.49 & 45.71 \\
 & 20 & 43.35 & 48.22 & 50.61 & \textbf{47.11} & 42.48 & 40.70 & 40.82 & 42.49 & 44.99 \\
\midrule
DASH~\cite{dash} & 30 & 43.84 & 48.54 & 50.20 & 44.83 & 44.59 & 42.79 & 39.70 & 40.34 & 44.83 \\
 & 20 & 43.84 & 47.25 & 49.80 & 44.07 & 42.74 & 40.70 & 39.33 & 42.49 & 44.10 \\
\midrule
OmniDrop & 30 & \textbf{44.33} & \textbf{52.43} &\textbf{51.22} & 46.35 & 43.80 &\textbf{45.12} &\textbf{45.32} & \textbf{45.06} &\textbf{46.78} \\
 & 20 & 43.60 & 50.49 & 51.63 & 45.59 & \textbf{45.12} & 43.26 & 44.94 & 43.78 & 45.06 \\
\bottomrule
\end{tabular}
}
\end{table}

\begin{table}[ht]
\centering
\caption{Performance comparison of various methods on the AVUT~\cite{avut} benchmark.}
\label{tab:supp_avut_app}
\small
\begin{tabular}{lcccccccc}
\toprule
Method & Ret. Ratio (\%) & EL & OR & OM & IE & CC & CM & Avg. \\
\midrule
\textit{Qwen2.5-Omni-7B} & 100 & 37.65 & 69.77 & 60.71 & 85.28 & 44.07 & 67.39  & 65.17 \\
\midrule
OmniZip~\cite{omnizip} & 30 & 35.88 & \textbf{68.49} & 55.10 & 83.13 & 39.83 & 63.07 & 61.76 \\
 & 20 & 36.47 & 65.59 & 55.10 & 83.44 & 39.83 & 61.87 & 61.07 \\
\midrule
DASH~\cite{dash} & 30 & \textbf{37.06} & 64.95 & 54.08 & 86.20 & 40.68 & 60.19 & 60.96 \\
 & 20 & \textbf{37.06} & 66.24 & 52.04 & 86.20 & 40.68 & 57.55 & 60.09 \\
\midrule
OmniDrop & 30 & \textbf{37.06} & \textbf{68.49} & 57.14 & 85.89 & 40.68 & \textbf{67.63} & \textbf{64.01} \\
 & 20 & 35.88 & 67.52 & \textbf{57.91} & \textbf{86.81} & \textbf{41.53} & 65.71 & 63.67 \\
\midrule[1.5pt]
\textit{Qwen2.5-Omni-3B} & 100 & 31.76 & 65.59 & 58.67 & 85.28 & 43.22 & 63.55 & 62.40 \\
\midrule
OmniZip~\cite{omnizip} & 30 & 32.94 & 62.70 & 53.83 & 84.36 & 44.07 & \textbf{61.39} & 60.27 \\
 & 20 & 32.94 & 62.70 & 50.77 & 83.13 & 43.22 & \textbf{61.39} & 59.28 \\
\midrule
DASH~\cite{dash} & 30 & 34.12 & 62.70 & 54.59 & \textbf{84.97} & 41.53 & 60.19 & 60.21 \\
 & 20 & \textbf{34.71} & 61.41 & 51.79 & 84.66 & 39.83 & 57.07 & 58.48 \\
\midrule
OmniDrop & 30 & 31.18 & \textbf{65.27} & \textbf{58.16} & 84.36 & \textbf{44.07} & \textbf{61.39} & \textbf{61.53} \\
 & 20 & 31.76 & 63.67 & 56.89 & 83.74 & 43.22 & 60.19 & 60.55 \\
\bottomrule
\end{tabular}
\end{table}

As shown in Tab.~\ref{tab:supp_ws_app}, OmniDrop shows robust performance across a wide range of WorldSense domains. For Qwen2.5-Omni-7B, OmniDrop achieves the best average scores at both 30\% and 20\% retained ratios, while also obtaining top or highly competitive results in multiple domains such as Music, Culture \& Politics, Sports, Film \& TV, and Games. This suggests that the gain of OmniDrop does not come from a narrow improvement in a single category, but from consistently preserving informative tokens across diverse domains. This tendency is also clear in the Qwen2.5-Omni-3B setting. At the 30\% retained ratio, OmniDrop achieves the best scores in most domains, except for Daily Life and Film \& TV. Even at the more aggressive 20\% retained ratio, OmniDrop remains competitive across categories and achieves strong scores in Technology \& Science, Film \& TV, Performance, and Games. 

Tab.~\ref{tab:supp_avut_app} presents detailed sub-task results on the AVUT benchmark, covering event localization (EL), object matching (OM), OCR matching (OR), information extraction (IE), content counting (CC), and character matching (CM). OmniDrop consistently achieves strong and balanced performance across these diverse sub-tasks. For Qwen2.5-Omni-7B, OmniDrop obtains the best average scores at both 30\% and 20\% retained ratios. In particular, it achieves the best performance on all sub-tasks, demonstrating that the method preserves audiovisual cues required for different types of audio-centric video understanding tasks. A similar trend is observed for Qwen2.5-Omni-3B. At the 30\% retained ratio, OmniDrop performs best on most of the sub-tasks. Although OmniDrop shows lower accuracy than others on EL under the 3B setting, it still achieves performance comparable to the original model. These results further confirm that OmniDrop offers a robust and balanced pruning strategy, maintaining task-relevant information across diverse tasks under aggressive compression.

\section{Analysis of omni-modal processing in Qwen2.5-Omni-3B}

\begin{figure}
    \centering
    \includegraphics[width=\textwidth]{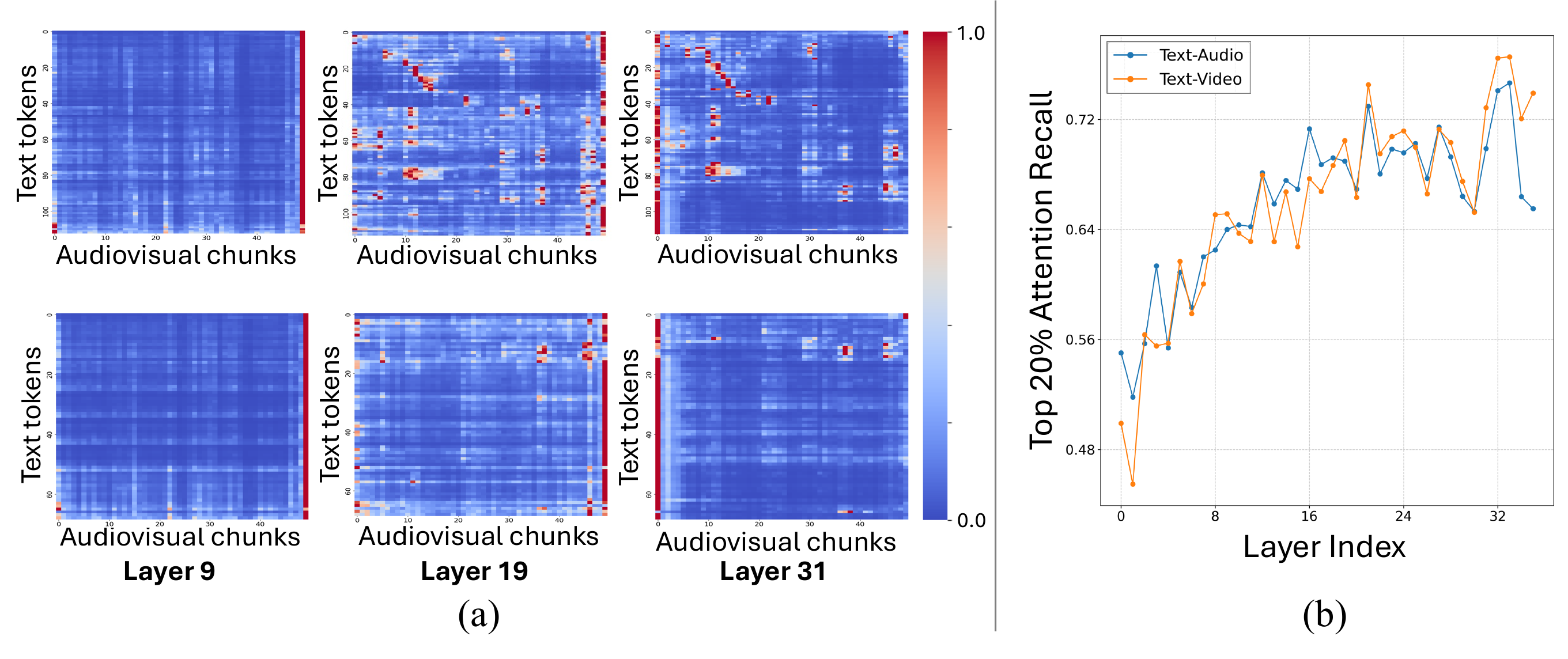}
    \caption{(a) Layer-wise text-to-audiovisual attention scores, averaged within each audiovisual chunk and normalized for better visualization. We show representative decoder layers, i.e., Layers 9, 19, and 31. The first and second rows correspond to the queries ``\textit{Who said in the video that `...'?}'' and ``\textit{In which part of the video does the man with ... appear?}'', respectively.
(b) Top-20\% attention recall
of text-to-audio and text-to-video attention scores across decoder layers.}
    \label{fig:layerwise_3B}
\end{figure}

While the primary analysis in Sec.~\ref{sec:analysis} focuses on Qwen2.5-Omni-7B, we further investigate the 3B variant to verify whether the same trend holds across model scales. In Fig.~\ref{fig:layerwise_3B}(a), we visualize text-to-audiovisual attention patterns using the same sample videos and queries as in Sec.~\ref{sec:analysis}.

Similar to the 7B model, the attention maps of the 3B model also highlight audiovisual chunks corresponding to the correct answers for both queries.
This indicates that the 3B model can naturally distinguish query-specific evidence after sufficient decoder-layer processing.

We further report the top-20\% attention coverage of text-to-other-modality interactions, following the same setting as the 7B analysis. As shown in Fig.~\ref{fig:layerwise_3B}(b), the coverage remains below 0.6 in the early layers, then gradually increases and reaches peaks above 0.7 from the middle layers onward. This suggests that attention becomes increasingly concentrated on a smaller set of query-relevant audiovisual tokens.

Overall, the 3B model shows a trend consistent with the 7B model, supporting our assumption that informative audiovisual tokens emerge progressively across decoder layers. These findings further justify applying our layer-wise adaptive pruning strategy to different model scales.

\section{Determining pruning ratios for a fixed token budget}
We implement a sigmoid-scheduled pruning rate with a high steepness parameter ($k=20$), creating a quasi-step transition that bisects the network into preservation and aggressive compression phases. This approach ensures a clear functional separation between early-layer omni-modal fusion and late-layer computational efficiency. 
To facilitate fair comparison with prior methods that operate under a fixed token budget, we calibrate our sigmoid schedule to retain 30\% of tokens on average. Thus, we derive the $p_{\text{final}}$ parameter necessary to maintain a global token budget of 30\% for the Qwen2.5-Omni-7B architecture ($L=28$). This derivation is conducted under the specific constraints of $t_{\text{mid}}=0.5$ and $p_{\text{init}}=0$.

\subsection{Cumulative token retained ratio dynamics}
Given that pruning is applied cumulatively across layers, the retained ratio of audio-video tokens $r_l$ just before being fed into layer $l$ evolves as a geometric decay process:

$$r_{l+1} = r_{l} \cdot (1 - p_l).$$

Starting with an initial retained ratio $r_0= 0.45$—a value established through a preliminary intra-modality pruning stage to remove intra-modal redundancies—the objective is to reach a target mean retained ratio $\bar{R} = \frac{1}{L}\sum_{l=0}^{L-1} r_l = 0.30$ over $L=28$ layers.

\subsection{Pruning ratio optimization for target retained ratio}
Under the bisection strategy ($p_{\text{init}} = 0, t_{\text{mid}} = 0.5$), the model maintains the initial retained ratio for the first half of the layers ($l \in [0, 13]$) and initiates exponential decay in the second half ($l \in [14, 27]$). 
The mean retained ratio $\bar{r}$ is approximated by the arithmetic mean of the two phases:

$$\bar{R} \approx \frac{\bar{r}_{\text{phase1}} + \bar{r}_{\text{phase2}}}{2}$$

1.  Phase 1 (Preservation): Since $p_{\text{init}} = 0$, the mean retained ratio is constant: $\bar{r}_{\text{phase1}} = r_0 = 0.45$.

2.  Phase 2 (Aggressive compression): To satisfy $\bar{R} = 0.30$, the required mean retained ratio for Phase 2 is:
    $$\bar{r}_{\text{phase2}} = 2\bar{R} - \bar{r}_{\text{phase1}} = 2(0.30) - 0.45 = 0.15$$
    
Using a linear approximation of the geometric decay and Taylor expansion, the mean retained ratio of Phase 2 corresponds approximately to the ratio at midpoint of the Phase 2 (\textit{i.e.} Layer $21$, the 7-th layer in the Phase 2). Thus:
    $$r_0 \cdot (1 - p_{\text{final}})^7 \approx \bar{r}_{\text{phase2}}$$
    $$(1 - p_{\text{final}})^7 \approx \frac{0.15}{0.45} = 0.333$$
    $$p_{\text{final}} \approx 1 - (0.333)^{1/7} \approx 0.146$$

While the theoretical derivation suggests $p_{\text{final}} \approx 0.146$ to converge at the target mean retained ratio of $30\%$, we adopt a more conservative setting of $p_{\text{final}} = 0.2$. This provides a safety margin to guarantee that the final average ratio remains strictly below the $0.3$ threshold across various input sequences. Furthermore, for scenarios requiring more extreme pruning, the initial pruning ratio $p_{\text{init}}$ can be increased above zero.

\clearpage
%%%%%%%%%%%%%%%%%%%%%%%%%%%%%%%%%%%%%%%%%%%%%%%%%%%%%%%%%%%%

\end{document}